\documentclass{article}
\usepackage{algorithm} 
\usepackage{algorithmicx}
\usepackage{algpseudocode}

\usepackage[preprint]{neurips_2026}

\usepackage[utf8]{inputenc} 
\usepackage[T1]{fontenc}    
\usepackage{hyperref}       
\usepackage{url}            
\usepackage{booktabs}       
\usepackage{amsfonts}       
\usepackage{nicefrac}       
\usepackage{microtype}      
\usepackage{xcolor}         
\usepackage{graphicx}
\usepackage{amsmath}
\usepackage{booktabs}
\usepackage{multirow}
\usepackage{tikz}
\usepackage{pgfplots}
\usepackage{pgfplotstable}
\usepgfplotslibrary{groupplots}
\pgfplotsset{compat=newest}
\usepackage{tabularx}
\usepackage{makecell}
\usepackage{enumitem}
\usepackage{adjustbox}
\usepackage{float}
\usepackage{wrapfig}
\usepackage{caption}
\usepackage{capt-of} 
\usepackage{amssymb}

\title{ATLAS: A Large-Scale Evaluation Benchmark for Adversarial LiDAR Perception}

\author{%
  Mellon M. Zhang\textsuperscript{*}
  \quad
  Siddhant Panse\textsuperscript{*}
  \quad
  Zimo Fan
  \quad
  Akshal Dhal
  \\
  \textbf{Rishit Sarkar}
  \quad
  \textbf{Glen Chou}
  \\[0.5em]
  Georgia Institute of Technology
  \\[0.5em]
  \texttt{\{meilongz, spanse30, zfan321, adhal8, rsarkar44, chou\}@gatech.edu}
}

\begin{document}

\maketitle
\def\thefootnote{*}\footnotetext{Equal contribution.}

\begin{abstract}
  Autonomous driving perception is typically evaluated on clean benchmark data, yet real-world deployment requires robustness to rare, structured, and potentially adversarial sensor anomalies. This gap is especially critical for LiDAR, where external actors can physically manipulate the sensing process to induce black-box perception failures without accessing the model. Existing LiDAR benchmarks provide little visibility into this failure mode. Prior adversarial LiDAR studies have largely centered on attack hardware, geometric and algorithmic defenses, and early-generation detectors, leaving the robustness of modern perception systems unexplored. To address this evaluation gap, we introduce ATLAS (Adversarial Temporal LiDAR Attack Suite), the first large-scale, physically grounded evaluation benchmark for LiDAR perception models under black-box sensor attacks, simulating the two primary attack modes -- point injection and point removal -- across real driving sequences. Evaluating a broad cross-section of current state-of-the-art LiDAR perception models, ATLAS reveals a surprising robustness asymmetry: models with stronger performance on standard benchmarks tend to better withstand removal attacks, yet are actually more vulnerable to injection attacks than weaker models. We trace this vulnerability to standard object database sampling augmentations, revealing how current training practices can induce architecture-agnostic robustness failures, and study initial directions for mitigating both attack modes. We release the ATLAS generation code to support extensible, reproducible evaluations as attack capabilities evolve, helping make black-box sensor robustness an explicit consideration in future LiDAR perception development.
\end{abstract}

\section{Introduction}
\label{sec:intro}
Accurate 3D perception is a core requirement for autonomous driving, where vehicles must localize, recognize, and react to surrounding agents in dynamic environments. Modern perception stacks combine cameras, LiDAR, radar, temporal tracking, and map priors for reliability under diverse conditions. Among these, LiDAR remains a central source of metric 3D geometry: by actively scanning with laser pulses and measuring returns, it provides precise 3D structures difficult to recover from RGB imagery alone, making LiDAR-based 3D object detection a major benchmark task.

However, the same active sensing process that makes LiDAR valuable also creates a distinct physical attack surface. Unlike passive cameras, LiDAR sensors construct point clouds through an externally observable measurement process that an adversary can exploit without accessing the detector architecture, weights, or training data. In a \emph{point injection} attack, carefully timed external laser signals create phantom geometry resembling real objects; in a \emph{point removal} attack, legitimate returns are suppressed, causing real structure to disappear. These attacks are black-box and sensor-level, corrupting the LiDAR stream before perception begins.

Current LiDAR benchmarks evaluate performance under benign conditions, capturing variation in weather, lighting, and traffic, but do not test whether perception models remain reliable when sensing is maliciously manipulated. This leaves a critical gap: as autonomous vehicles move toward public deployment, robustness to adversarial sensor corruption must be measurable during development. Moreover, robustness cannot be assumed from clean-data accuracy, as stronger detectors, temporal aggregation, and multimodal fusion may reject or amplify corrupted geometric evidence in unpredictable ways.

We introduce \textbf{ATLAS} (\textbf{A}dversarial \textbf{T}emporal \textbf{Li}DAR \textbf{A}ttack \textbf{S}uite), a large-scale, physically grounded benchmark for evaluating black-box LiDAR spoofing robustness in real driving sequences. Built on the Waymo Open Dataset, ATLAS simulates point injection and point removal with attack budgets aligned with emerging hardware capabilities, enabling systematic evaluation of how modern 3D object detectors behave when the LiDAR stream is corrupted before perception.

Using ATLAS, we conduct a broad empirical audit of modern LiDAR perception systems, including single-frame, temporal, streaming, and camera-LiDAR fusion detectors across convolutional, transformer, and state-space architectures. Our central finding is that progress on clean benchmarks does not reliably translate to spoofing robustness: stronger models can better tolerate point removal but become more susceptible to point injection, revealing a concerning robustness asymmetry that motivates adversarial sensor-level evaluation as a first-class criterion.

Our contributions are as follows:
\begin{enumerate}[nosep]
    \vspace{-0.5em}
    \item We introduce \textbf{ATLAS}, a large-scale benchmark for evaluating black-box adversarial LiDAR spoofing under physically grounded point injection and removal attacks in real sequences.
    \item We conduct a comprehensive robustness audit of 3D object detectors across single-frame, temporal, streaming, camera-LiDAR fusion, and diverse architectural families.
    \item We identify a systematic robustness asymmetry: stronger clean-benchmark models can partially recover from point removal, but often become more vulnerable to point injection.
    \item We release ATLAS to support extensible evaluation of LiDAR perception robustness under evolving sensor-level attack capabilities.
\end{enumerate}
\vspace{-1em}
\section{Related Work}
\vspace{-1em}
\label{sec:related_works}

\noindent
\begin{minipage}[t]{0.50\textwidth}
    \vspace{0pt}
    \noindent \textbf{LiDAR Perception Benchmarks}.
    Large-scale driving datasets have been the primary basis of LiDAR-based 3D object detection research. KITTI \cite{kitti} established the first multimodal benchmark, with subsequent datasets increasing scale and diversity \cite{nuScenes, zod, RoScenes, V2X-Real}: nuScenes introduced a full 360\textdegree\ sensor suite, Argoverse 2 \cite{Argoverse} focused on HD maps, and Waymo Open \cite{Waymo} provided over 12 million 3D annotations. However, these datasets largely assume a trustworthy sensing process and do not account for malicious manipulation of LiDAR measurements. Existing robustness benchmarks such as KITTI-C, nuScenes-C, and Waymo-C \cite{CorruptionRobustnessBenchmarksC} primarily evaluate performance under natural corruptions including weather, sensor noise, and beam dropout, leaving adversarial sensing attacks largely unexplored (Fig.\ref{fig:teaser}). This missing evaluation setting motivates the development of ATLAS, a benchmark specifically designed to study adversarial robustness in LiDAR perception systems.
    \noindent \textbf{LiDAR Spoofing}.
LiDAR sensors are vulnerable to physical attacks that inject phantom objects or remove legitimate returns \cite{sun2020lidarblackbox, sato2023removal, jin2023pla}. Sun et al.\ 
\end{minipage}
\hfill
\begin{minipage}[t]{0.4\textwidth}
    \vspace{-3em}
    \centering
    \includegraphics[width=\linewidth]{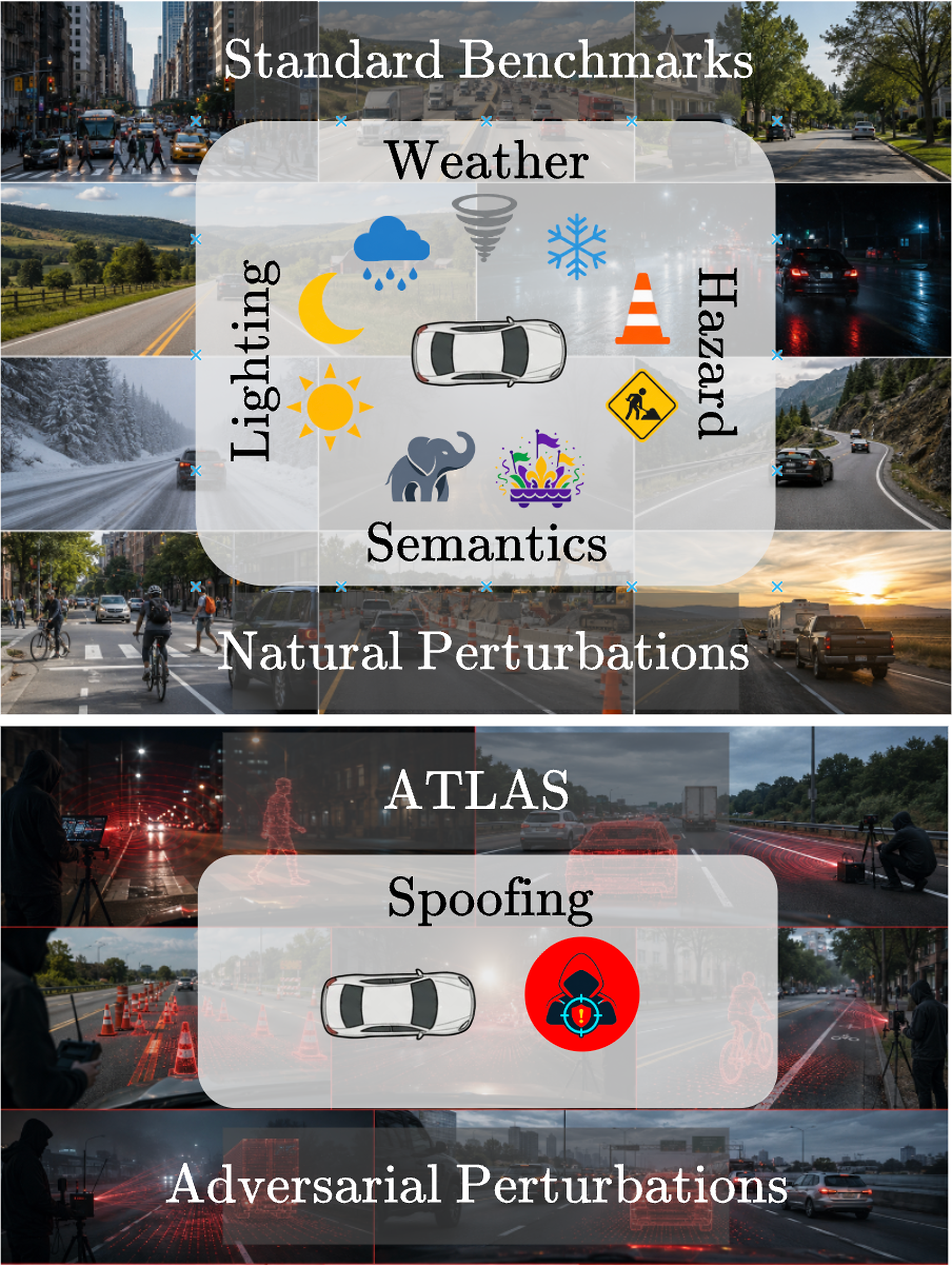}
    \captionof{figure}{\textbf{ATLAS fills a missing axis in LiDAR robustness evaluation.}
While existing benchmarks primarily study natural perturbations, ATLAS targets black-box, sensor-level spoofing attacks.}
    \label{fig:teaser}
\end{minipage}

\vspace{-0.5em}
  \cite{sun2020lidarblackbox} demonstrated feasibility with 140 structured injected points, PLA-LiDAR \cite{jin2023pla} scaled to 4200 points with 51--98\% false detection rates on PointPillars \cite{PointPillar} and SECOND \cite{SECOND}, PhantomLidar \cite{jin2025phantomlidar} injected 16,000 points via electromagnetic interference, and Sato et al.\ \cite{sato2025lidar} showed 1,000 wall-arranged points suffice to trigger emergency braking. Beyond injection, Sato et al.\ \cite{sato2023removal} demonstrated hiding real vehicles via the sensor's denoising mechanism, and their follow-up work \cite{sato2025lidar}, which grounds our removal simulation, introduced the adaptive high-frequency removal (A-HFR) attack, showing a 20\textdegree\ azimuthal sector suffices to eliminate all points from a vehicle at range. Cao et al.\ \cite{cao2023you} further confirm removal via relay and jamming. Despite these advances, evaluation protocols remain anchored to early single-frame detectors \cite{PointPillar, SECOND, CenterPoint, abraha2025convnet, tu2020physicallyrealizableadversarialexamples, Cao_2019, kobayashi2025invisible}, leaving modern architectures and temporal methods unevaluated, with no open-source benchmark unifying these attack models.
\vspace{-1em}
\section{Adversarial Temporal LiDAR Attack Suite (ATLAS)}
\vspace{-1em}

\begin{figure}[!h]
    \centering
    \includegraphics[width=\linewidth]{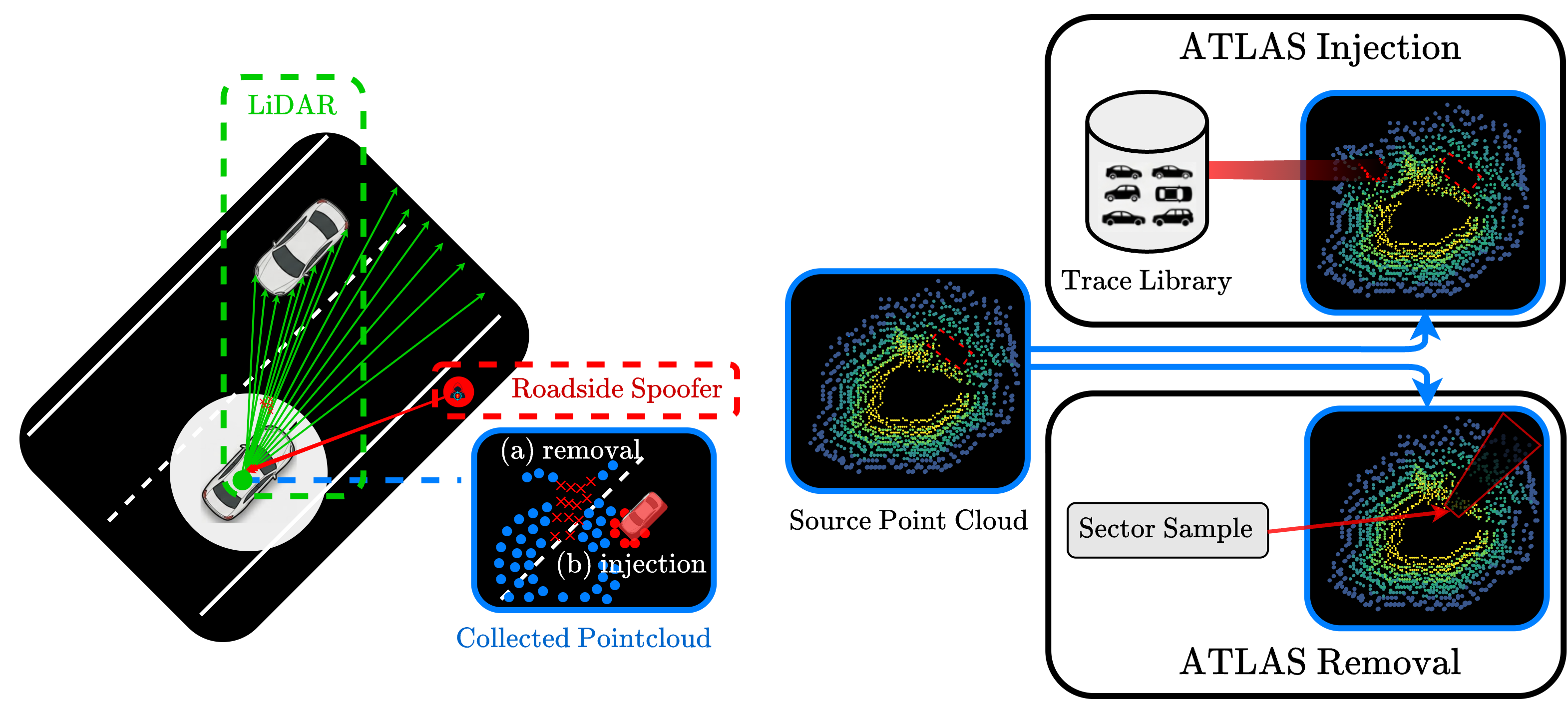}
    \caption{Atlas simulates the main attack modes. Point injection attacks inject spurious objects to induce sudden driving control change. Point removal attacks interfere with the scanning process to hide real objects from sensor readings.}
    \label{fig:attack_scope}
    \vspace{-2em}
\end{figure}

\subsection{Threat Model} 
\vspace{-1em}
We consider the physically realizable roadside spoofer threat model shown in Figure \ref{fig:attack_scope} \cite{sato2025lidar}\cite{jin2023pla}, where an attacker near the ego vehicle emits synchronized laser signals to inject or remove LiDAR points, creating false measurements. Prior work has shown that such attacks can inject thousands of points per frame \cite{jin2023pla, jin2025phantomlidar}, sufficient to form object-like structures \cite{sun2020lidarblackbox}. This threat model is practical and requires no access to model parameters, relying only on publicly available LiDAR scan patterns \cite{nuScenes, Waymo, Argoverse, kitti}. Since the attack directly manipulates the physical sensing process, it is difficult to detect with standard cybersecurity defenses. We focus on the two main attack modes: point injection \cite{jin2023pla} and point removal \cite{sato2025lidar}.

\vspace{-1em}
\subsection{Benchmark Scope}\label{sec:benchmark_scope}
\vspace{-1em}
We construct a large-scale benchmark emulating modern physical LiDAR spoofing on multi-frame detection pipelines, built on the Waymo Open Dataset (WOD) \cite{Waymo} for realistic driving scenarios with diverse traffic and sensor conditions.
For point injection, we vary point budget and phantom-object position; for removal, we vary azimuthal attack width, which determines the expected point removal rate. Both families share the same temporal attack schedule (Sec. ~\ref{sec:temp_attack}), yielding twelve variants of the WOD validation split.
Altogether, ATLAS comprises $>460\text{k}$ LiDAR point clouds across $195$ diverse driving trajectories each approximately $20$ seconds long.

\noindent\textbf{Point injection point budget.\quad}
We vary injected point count across three regimes: \textit{easy} (100--200 points), aligned with early spoofing devices \cite{sun2020lidarblackbox}; \textit{medium} (300--500 points), and \textit{hard} (500--1000 points), both representative of state-of-the-art systems such as PLA-LiDAR \cite{jin2023pla}, which can inject up to 4200  points of any arbitrary geometry.

\noindent\textbf{Phantom object position.\quad}
We consider two physically grounded modes: \textit{relative-position fixed}, where the injected object maintains a constant pose relative to the ego vehicle, simulating an adversary moving alongside it \cite{cao2023you}; and \textit{global-position fixed}, where the spoofed object is placed at a fixed world-frame location, simulating a static roadside attacker injecting a phantom parked vehicle as the ego approaches \cite{sun2020lidarblackbox, sato2023removal, sato2025lidar, wip}.

\noindent\textbf{Point removal rate.\quad}
We design the point removal attacks to reflect the capabilities of the adaptive high-frequency removal (A-HFR) attack from \cite{sato2025lidar}. A-HFR removes points by emitting high-frequency LiDAR pulses only when the sensor is scanning a target region, disrupting true returns so the object is not recorded. Sato et al. demonstrate that the target azimuthal sector width directly constrains removal efficacy. We construct six datasets of varying target azimuthal sector widths (10\textdegree---60\textdegree) and adopt empirically-validated point removal rates (PRR) from \cite{sato2025lidar}, restated in Table \ref{tab:prr}.
\begin{table}[h]
\begin{center}
\caption{Target removal azimuth vs Point removal rate (PRR)}
\begin{tabular}{c|cccccc}
Azimuth & 10° & 20° & 30° & 40° & 50° & 60° \\
\hline
PRR & 97\% & 100\% & 99\% & 98\% & 90\% & 90\%
\end{tabular}
\label{tab:prr}
\end{center}
\vspace{-2em}
\end{table}

\noindent\textbf{Attack timing.\quad}
We model the temporal evolution of spoofing attacks by considering both their onset and accumulation. At attack onset, the multi-frame detectors operate on clean historical memory, allowing us to isolate the immediate impact of injected points. As the attack persists, spoofed observations are incorporated into the model’s memory, enabling us to study how errors accumulate and propagate over time within multi-frame pipelines.
\vspace{-1em}
\subsection{Temporal attack schedule}
\label{sec:temp_attack}
\vspace{-0.8em}
To model sustained attack behavior, we schedule spoofing in alternating temporal windows within each driving segment of the WOD validation split. We partition each sequence into 32-frame blocks and alternate between clean and spoofed intervals. Let $s_i \in \{0,1\}$ denote whether frame $i$ is spoofed. We initialize with a clean frame ($s_0 = 0$), and for $i \geq 1$ define the attack schedule in Eq. \ref{eq:schedule}.
\begin{equation}
\label{eq:schedule}
s_i = \mathbf{1}\!\left[\lfloor i/32 \rfloor \bmod 2 = 1\right].
\end{equation}

This attack schedule is applied to both the point injection and removal benchmarks. For point injection, this produces a repeating pattern in which a phantom vehicle appears, persists for 32 frames ($\approx$ 3.2 seconds), disappears, and reappears after an equally long clean interval. For point removal, it defines a window in which several frames contain point clouds where 90-100\% of points within an azimuthal sector are removed. Unlike point injection, the attack is not guaranteed to persist across each frame within the 32-frame window, due to probabilistic modeling of simulated spoofing device functionality (see \textbf{\hyperref[para:stochastic_point_removal]{Stochastic point removal})}.

We use a 32-frame horizon, matching both the practical capability of existing LiDAR spoofing systems to sustain multi-second attacks \cite{wip} and the upper range of temporal context used by modern multi-frame detectors (4--32 frames). This setting allows us to evaluate how adversarial signals interact with the full memory capacity of temporal perception models.

\vspace{-1em}
\subsection{Point Injection Benchmark Construction}
\vspace{-0.8em}
\noindent \textbf{Trace library.}\label{para:trace_library}
Rather than inserting synthetic points, we simulate sensor-level injection to approximate physical attacks \cite{sun2020lidarblackbox, jin2023pla}. We extract annotated vehicles from Waymo Open \cite{Waymo} and aggregate their LiDAR returns in a canonical vehicle frame, preserving range-dependent sparsity, occlusion patterns, and intensity distributions. Replaying these traces into new scenes synthesizes physically plausible phantom vehicles, closely following the record-based spoofing of PLA-LiDAR \cite{jin2023pla}, which replays recorded LiDAR observations via synchronized laser injection to induce false detections.

To generate an attack instance, we sample a vehicle trace and target placement relative to the ego vehicle: longitudinal offset $x \sim \mathcal{U}(15\text{m}, 20\text{m})$ and lateral offset $y \sim \mathcal{U}(-3\text{m}, 3\text{m})$. The trace is centered on its annotated box, perturbed with random yaw $\psi \sim \mathcal{U}(-0.2, 0.2)$, translated to the sampled location, ground-aligned, and assigned range-consistent intensity and elongation statistics.

\noindent\textbf{Modeling adversary position}
In the \textit{relative-position fixed} setting, the ghost vehicle is initialized at a sampled ego-relative offset $(x_0, y_0)$ (see \hyperref[para:trace_library]{\textbf{Trace Library}}) and maintains this offset throughout the spoof window. As a result, the injected object remains at a constant bearing and distance relative to the ego vehicle, mimicking an adversary that tracks or moves alongside it.

In the \textit{global-position fixed} setting, the spoofed object is anchored in the world frame and does not move as the ego vehicle approaches it. We define an \emph{anchor frame} as the final frame of each 32-frame spoof block. The object is first placed at the sampled ego-relative offset in this anchor frame and then transformed into global coordinates, as illustrated in Eq. \ref{eq:global_place}.
\begin{equation}
\mathbf{p}^{\mathrm{world}} = \mathbf{R}_{\mathrm{anc}} \mathbf{p}^{\mathrm{ego}}_{\mathrm{anc}} + \mathbf{t}_{\mathrm{anc}}
\label{eq:global_place}
\end{equation}
$\mathbf{R}_{\mathrm{anc}}$ and $\mathbf{t}_{\mathrm{anc}}$ denote the ego-to-world rotation and translation. For each frame $k$ within the spoof window, these world-frame points are projected back into the ego coordinate system, shown in Eq.\ref{eq:global_to_ego}.
\begin{equation}
\mathbf{p}^{\mathrm{ego}}_k = \mathbf{T}_k^{-1} \mathbf{p}^{\mathrm{world}}.
\label{eq:global_to_ego}
\end{equation}
$\mathbf{T}_k \in SE(3)$ is the ego-to-global transformation at frame $k$. As the ego vehicle moves, the ghost object therefore follows a natural approach-and-recede trajectory in the ego frame while remaining stationary in the world.

\noindent \textbf{Sensor-level spoof simulation via raycasting.} \label{para:sensor_level} To enforce physical realism, we explicitly model occlusion through spherical raycasting in discretized sensor coordinates. Both scene and spoof points are first converted to spherical coordinates $(r, \theta, \phi)$ and binned into an $(M \times N)$ grid indexed by azimuth and beam.  Azimuth is discretized into $M = \lceil 360^\circ / 0.1358^\circ \rceil \approx 2651$ bins, matching Waymo's \cite{Waymo} native angular resolution, while elevation is quantized to the nearest of $N=64$ physical beam inclinations extracted from the sensor specification. This discretization maps the continuous raycasting problem into a finite set of independent rays, each identified by a unique (azimuth bin, beam index) pair. 

For each spoofed point at range $r_{\text{spoof}}$, we apply range-based visibility rules relative to scene returns $r_i$: (1) if $r_{\text{spoof}}$ lies behind all scene returns, the point is discarded; (2) if $r_{\text{spoof}}$ is the nearest return, it becomes the primary echo and occludes all scene returns; (3) if $r_i < r_{\text{spoof}} < r_j$, it occludes all farther returns $\{r_k: r_k \geq r_j\}$. In case (2), we retain the nearest return behind the spoof as a secondary pulse with probability $p_{\text{dual}}=0.01$ to approximate dual-return behavior \cite{li2024lidar_returns}. 

The visible spoof and scene points are merged to form the final point cloud for each frame. All injected points are processed identically to real measurements, meaning that they enter the pipeline prior to voxelization. When a spoofed object is present in the scene, its ground-truth bounding box is transformed into the current ego frame and recorded as $g \in \mathbb{R}^8$ (comprising $[x, y, z, l, w, h, \theta, c]$ where $\theta$ is heading and $c$ is the object class). The bounding boxes of the spoofed objects are recorded for downstream attack success rate computation (Sec. ~\ref{sec:asr}).

The full process of generating the point injection benchmark is summarized with Algorithm ~\ref{alg:injection} in Appendix \ref{app:algs}.

\vspace{-1em}
\subsection{Point Removal Benchmark Construction}
\vspace{-0.8em}
\noindent \textbf{Target selection and tracking} Unlike the phantom point injection attack, the point removal attack removes real LiDAR points from an angular sector aimed at a target object. For each 32-frame attack window, we select an initial vehicle target from the Waymo ground-truth objects. We then track the initial target through the attack window using a simple velocity-guided association rule. Given the previous box center $\mathbf{c}_{k-1}$ and planar velocity $(v_x, v_y)$, we use the relation in Eq. \ref{eq:removal_tracking} to predict the next-frame target center.
\begin{equation}
\hat{\mathbf{c}}_k =
\mathbf{c}_{k-1}
+
\Delta t
\begin{bmatrix}
v_x & v_y & 0
\end{bmatrix}^{\top}
\label{eq:removal_tracking}
\end{equation}
We choose $\Delta t = 0.1\text{s}$ to be consistent with Waymo's 10 Hz LiDAR \cite{Waymo}. The target in frame $k$ is then chosen as the nearest same-class ground truth box to $\hat{\mathbf{c}}_k$, subject to a maximum displacement limit and a bounding-box volume consistency check. If no valid match is found, the target is considered lost and no removal attack is applied for that frame.

\noindent \textbf{Removal sector construction.} For each frame where the object targeted for removal is successfully tracked, we aim the removal sector at the center of the tracked target box. Let $\mathbf{p}_i=(x_i,y_i,z_i)$ denote a LiDAR point and let $(\theta_i,\phi_i)$ be its azimuth and elevation. Similarly, let $(\theta_k^{\star},\phi_k^{\star})$ be the azimuth and elevation of the tracked target center in frame $k$. The removal  sector is defined in Eq. \ref{eq:ahfr_sector}
\begin{equation}
\mathcal{S}_k =
\left\{
i:
\left|\mathrm{wrap}(\theta_i-\theta_k^{\star})\right|
\leq \frac{\alpha}{2},
\quad
\left|\phi_i-\phi_k^{\star}\right|
\leq \frac{\beta}{2}
\right\},
\label{eq:ahfr_sector}
\end{equation}
We define $\alpha$ as the azimuth attack width and $\beta$ as the elevation width. Following \cite{sato2025lidar}, we vary $\alpha$ but fix $\beta = 16^\circ$. The azimuth width $\alpha$ determines the point-removal probability $p_{\mathrm{remove}}$ according to the empirical relation shown in Table \ref{tab:prr}.

\noindent \textbf{Stochastic point removal.} \label{para:stochastic_point_removal}
We simulate point removal deployment as a stochastic removal conditioned on ego-to-target distance. Sato et al. \cite{sato2025lidar} report distance-binned A-HFR success rate and attribute the attack not firing to missed weak-synchronization events. The attack is triggered only when the optical receiver detects the victim LiDAR pulse, so scans with missed LiDAR pulses fail to launch the attack. Accordingly, after selecting a target object to hide, we sample whether the attack fires using the empirical distance-dependent success probabilities averaged over vehicle speeds (Table XI in \cite{sato2025lidar}). This is summarized by the distribution in Eq. \ref{eq:fireprob}.
\begin{equation}
\label{eq:fireprob}
    a_k \sim \mathrm{Bernoulli}(p_{\text{dist}}(r_k))
\end{equation}
The firing probability is dependent on $p_{\text{dist}}(r_k)$, the distance-dependent firing probability and $r_k$,  the distance between the ego LiDAR and target object. If $a_k=1$, each point within $\mathcal{S}_k$ is removed with probability $p_{\text{remove}}$. Points outside of $\mathcal{S}_k$ are retained. This produces a structured angular deletion pattern rather than random point dropout across the full scene. 

For point removal, $g \in \mathbb{R}^8$ denotes the real tracked target box that the adversary attempts to hide. We store $g$ only for frames in which the target is successfully tracked and the firing gate is active; otherwise $g$ is set to null. This field is stored separately from the standard ground-truth boxes and is used only for attack success evaluation (See 
Sec. ~\ref{sec:asr}).

The full process of generating the point removal benchmark is summarized with Algorithm ~\ref{alg:removal} found in Appendix [\ref{app:algs}].
\vspace{-1em}
\section{Evaluation Setup}
\vspace{-1em}

We evaluate the detectors from Table \ref{tab:model_summary} on ATLAS. We train all models on 8 NVIDIA L40S GPUs and run each evaluaton on 1 NVIDIA H100 GPU. For an in-depth summary of baseline functionality, refer to Appendix ~\ref{app:baselines}. For detector-specific evaluation details, refer to Appendix~\ref{app:eval_implementation}.

\begin{table*}[t]
\centering
\small
\setlength{\tabcolsep}{4pt}
\renewcommand{\arraystretch}{1.05}
\caption{Summary of evaluated perception models.}
\label{tab:model_summary}
\resizebox{\textwidth}{!}{
\begin{tabular}{l c c c c c c}
\toprule
\textbf{Model} &
\textbf{Conf.} &
\textbf{Type} &
\textbf{Arch.} &
\textbf{Frames} &
\textbf{L2 mAP/mAPH} &
\textbf{Rank} \\
\midrule
CenterPoint~\cite{CenterPoint}        & CVPR'21 & Single-Frame & Conv.      & 1  & 65.7/63.3 & 9 \\
CenterFormer~\cite{CenterFormer}      & ECCV'22 & Single-Frame & Trans.     & 1  & 69.9/69.4 & 5 \\
DSVT~\cite{DSVT}                      & CVPR'23 & Single-Frame & Trans.     & 1  & 74.0/72.1 & 4 \\
PHiM~\cite{phim}                      & WACV'26 & Streaming    & Mamba      & 1  & 68.7/66.7 & 7 \\
LoGoNet~\cite{logonet}                & CVPR'23 & Cam-LiDAR    & Conv+Attn  & 1  & 68.6/65.9 & 8 \\
CenterPoint-4f~\cite{CenterPoint}     & CVPR'21 & Multi-Frame  & Conv.      & 4  & 69.1/67.6 & 6 \\
MSF-4f~\cite{He_2023_CVPR}            & CVPR'23 & Multi-Frame  & Conv+Attn  & 4  & 74.7/73.4 & 2 \\
MSF-8f~\cite{He_2023_CVPR}            & CVPR'23 & Multi-Frame  & Conv+Attn  & 8  & 74.4/73.0 & 3 \\
PTT~\cite{huang2024ptt}               & CVPR'24 & Multi-Frame  & Conv+Attn  & 32 & 74.9/73.6 & 1 \\
\bottomrule
\end{tabular}
}
\vspace{-1em}
\end{table*}

\noindent \textbf{Evaluation Metric.}\label{sec:asr}
We report attack success rate (ASR) to quantify spoofing effectiveness. For injection, an attack succeeds if any predicted box overlaps the phantom ground-truth box $g$ with IoU $\geq 0.7$ \cite{Waymo}. For removal, an attack succeeds if no predicted box overlaps the target ground-truth box $g$ at the same threshold, indicating a missed detection.
The ASR for both cases is then defined as the fraction of spoofed targets for which a successful false positive/negative is produced:
{\setlength{\abovedisplayskip}{0.5pt}
\setlength{\belowdisplayskip}{0.5pt}
\begin{equation}
\text{ASR} = \frac{\# \text{ of successful spoofed targets}}{\# \text{ of total spoofed targets}}.
\end{equation}}%
\pgfplotsset{compat=newest}
\begin{figure}[t]
\centering
\begin{minipage}{0.62\linewidth}
\centering
\resizebox{\linewidth}{!}{
\begin{tikzpicture}
        \begin{groupplot}[
            group style={
                group size=1 by 1,
            },
            width=9cm,
            height=8cm,
            xlabel={\textbf{Injection Robustness Rank}},
            ylabel={\textbf{Removal Robustness Rank}},
            xmin=0.0, xmax=9.8,
            ymin=0.0, ymax=9.3,
            clip=false,
            grid=none,
            axis lines=left,
            tick label style={font=\large},
            label style={font=\large},
            title style={font=\bfseries\Large},
            colormap={viridis}{
                rgb255=(68,1,84)
                rgb255=(59,82,139)
                rgb255=(33,145,140)
                rgb255=(94,201,98)
                rgb255=(253,231,37)
            },
            colorbar,
            colorbar style={
                xshift=0.5cm,
                ylabel={\textbf{Clean Performance Rank}},
                ylabel style={
                    font=\large,
                    at={(2.0,0.5)},
                    anchor=center,
                    rotate=0,
                },
                tick label style={font=\large},
                ymin=1, ymax=9,
                ytick={1,2,3,4,5,6,7,8,9},
            },
            point meta min=1,
            point meta max=9,
            scatter/use mapped color={
                draw=black,
                fill=mapped color,
            },
        ]

        \nextgroupplot[
            title={\textbf{Robustness Correlation}},
        ]

        \addplot[
            thick,
            dashed,
            color=cyan!70!blue,
            no marks,
            domain=1.0:9.5,
            samples=100,
        ] {-0.9596*x + 9.7980};

        \addplot[
            scatter,
            only marks,
            mark=*,
            mark size=5pt,
            mark options={draw=black, line width=0.8pt},
            scatter src=explicit,
        ] coordinates {
            (1.667, 8.667) [9]   
            (3.500, 7.333) [5]   
            (4.667, 6.000) [4]   
            (2.333, 7.500) [7]   
            (3.500, 5.333) [8]   
            (5.500, 3.500) [6]   
            (7.167, 3.167) [2]   
            (9.000, 2.333) [3]   
            (7.667, 1.167) [1]   
        };

        \node[anchor=south west, font=\small, fill=white, fill opacity=0.75, text opacity=1, inner sep= 4pt]
        at (axis cs:1.667, 8.667) {CP~\cite{CenterPoint}};
        
        \node[anchor=south west, font=\small, fill=white, fill opacity=0.75, text opacity=1, inner sep=4pt]
        at (axis cs:3.500, 7.333) {CF~\cite{CenterFormer}};
        
        \node[anchor=south west, font=\small, fill=white, fill opacity=0.75, text opacity=1, inner sep=4pt]
        at (axis cs:4.667, 6.000) {DSVT~\cite{DSVT}};
        
        \node[anchor=north east, font=\small, fill=white, fill opacity=0.75, text opacity=1, inner sep=4pt]
        at (axis cs:2.333, 7.500) {PHiM~\cite{phim}};
        
        \node[anchor=north east, font=\small, fill=white, fill opacity=0.75, text opacity=1, inner sep=4pt]
        at (axis cs:3.500, 5.333) {LGN~\cite{logonet}};
        
        \node[anchor=south, font=\small, fill=white, fill opacity=0.75, text opacity=1, inner sep=5pt]
        at (axis cs:5.500, 3.500) {CP4F~\cite{CenterPoint}};
        
        \node[anchor=south west, font=\small, fill=white, fill opacity=0.75, text opacity=1, inner sep=4pt]
        at (axis cs:7.167, 3.167) {MSF4~\cite{He_2023_CVPR}};
        
        \node[anchor=east, font=\small, fill=white, fill opacity=0.75, text opacity=1, inner sep=4pt]
        at (axis cs:8.85, 2.333) {MSF8~\cite{He_2023_CVPR}};
        
        \node[anchor=north east, font=\small, fill=white, fill opacity=0.75, text opacity=1, inner sep=4pt]
        at (axis cs:7.667, 1.167) {PTT~\cite{huang2024ptt}};

        \end{groupplot}
\end{tikzpicture}}
\end{minipage}
\hfill
\begin{minipage}{0.34\linewidth}
\caption{\textbf{Negative correlation between injection and removal robustness.}
Models with higher injection robustness tend to exhibit lower removal robustness
($R^2 = 0.87$). Each point represents a model; color encodes clean detection
performance on the benchmark (higher is better). The dashed line shows the linear
regression fit. Injection and removal robustness ranks are computed by averaging
the per-setting ASR rankings across all injection and removal configurations,
respectively, where lower ASR indicates higher robustness.}
\label{fig:robustness-correlation}
\end{minipage}
\vspace{-2em}
\end{figure}
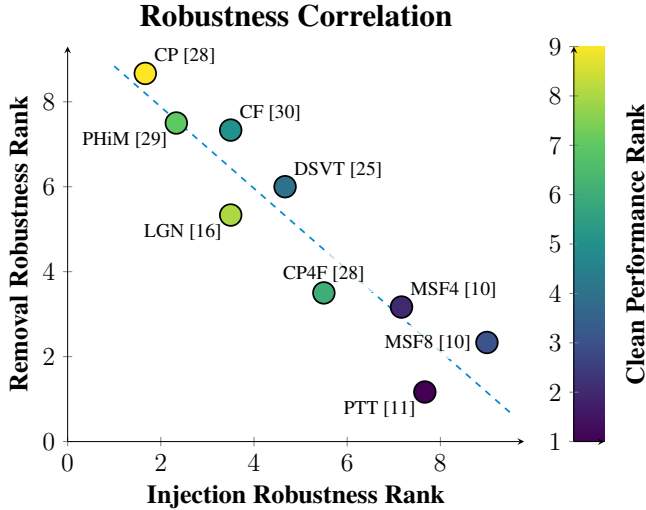
\vspace{-0.1em}
In addition to ASR, we analyze confidence scores to better understand model calibration under attack. We evaluate phantom object confidence for point injection and the confidence degradation of surviving objects for point removal.
\vspace{-1em}
\section{Results}\label{sec:results}
\vspace{-1em}
We structure our results section to highlight three critical findings from our evaluations on ATLAS.
\vspace{-0.5em}
\begin{enumerate}[nosep, topsep=0pt]
    \item Stronger clean benchmark performance improves robustness to point removal, but unexpectedly increases vulnerability to point injection (\hyperref[para:finding1]{\textbf{Finding 1}}). 
    \item Spoofed objects induce false high confidence across detectors (\hyperref[para:finding2]{\textbf{Finding 2}}).
    \item Temporal detectors rely heavily on current observations to make predictions  (\hyperref[para:finding3]{\textbf{Finding 3}}). 

\end{enumerate}%
For additional results, see Appendix\ref{app:extendedresults}.

\begin{figure}[!h]
    \centering
    \includegraphics[width=\linewidth]{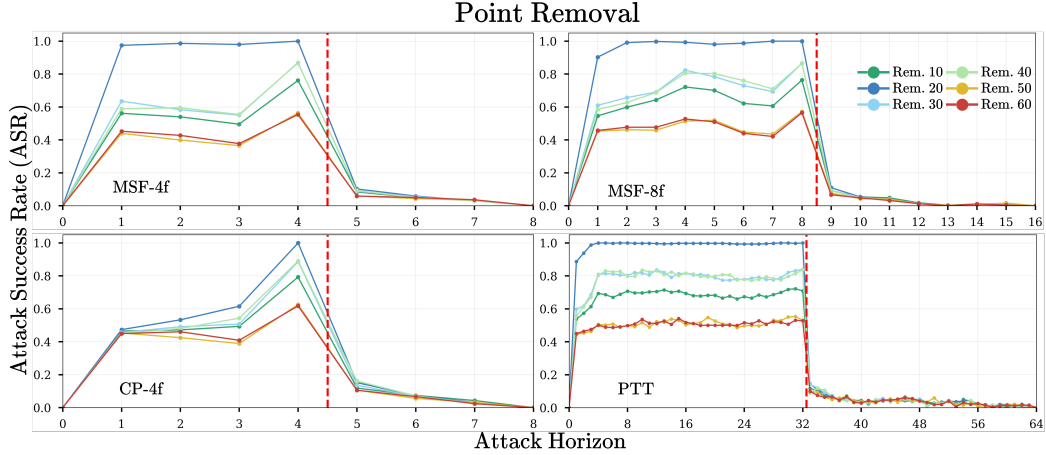}
    \vspace{-2em}
    \caption{\textbf{Decay of removal vulnerability.} The attack success rate peaks when the history is fully saturated before decaying monotonically as clean observations enter the temporal memory buffer.}
    \label{fig:attack_lifetime_removal}
    \vspace{-1em}
\end{figure}

\begin{figure}[!h]
    \centering
    \includegraphics[width=0.9\linewidth]{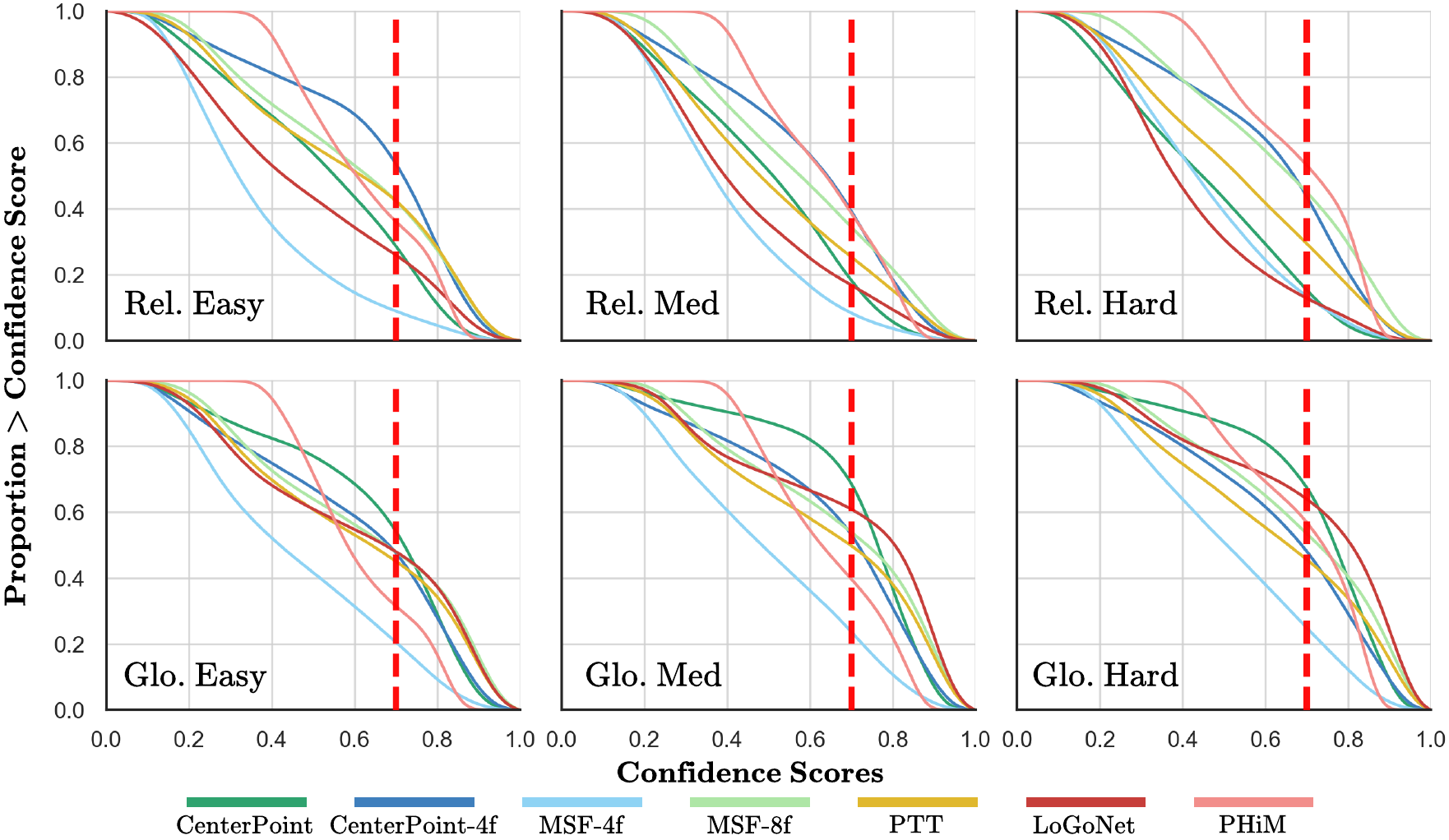}
    \vspace{-0.5em}
    \caption{\textbf{High proportion of high-confidence phantom proposals.} Across most models and point injection modes, at least 25\% of false positives are high in confidence (>0.7).}    \label{fig:confidence_injection}
    \vspace{-1em}
\end{figure}

\vspace{-1.1em} 
\paragraph{Finding 1.}\label{para:finding1}
Detection performance and adversarial robustness exhibit opposing trends across attack types, as illustrated by the negative correlation ($R^2=0.87$) in Figure ~\ref{fig:robustness-correlation}. Under point injection, single-frame detectors are the most robust: CenterPoint and PHiM yield the lowest ASR across all spoofing configurations (e.g., 0.32 and 0.33 under global-hard), while ranking among the weakest in clean detection performance (L2 mAP of 65.7 and 68.7, respectively - Table \ref{tab:model_summary}). Multi-frame temporal detectors are consistently more vulnerable. MSF-8f yields the highest injection ASR of any detector (0.54 under global-hard), followed by PTT (0.5). Notably, injection vulnerability is not strictly monotonic with clean performance: PTT's detection capability is marginally higher than that of MSF-8f and MSF-4f (reflected by a higher L2 mAP) yet is less susceptible to injection, suggesting that architectural differences in temporal aggregation, not model capacity alone, modulate injection vulnerability (Table \ref{tab:injection_results}). This trend reverses under point removal. PTT yields the lowest removal ASR of any detector (0.50 under removal-60), while CenterPoint and PHiM suffer ASRs of 0.77 at the same setting (Table ~\ref{tab:removal_results}). These results demonstrate that more performant models can recover from point removal attacks using stronger learned representations or temporal context, but are more vulnerable to point injection attacks. We analyze this phenomenon in further detail in Sec. \ref{sec:discussion}.
\vspace{-1.1em}
\paragraph{Finding 2.} \label{para:finding2}
Table \ref{tab:injection_results} shows that each point injection attack variant induces a non-trivial rate of phantom detections across all detectors. Figure \ref{fig:confidence_injection} reveals that these detections aren't all low-confidence artifacts easily filtered by score thresholding: across most detectors and point injection configurations, at least 20$\%$ of phantom detections receive confidence scores exceeding 0.7. PHiM is a notable outlier, with phantom confidence concentrated below 0.4 in the relative-fixed modes. As point density increases (easy vs medium vs hard), the confidence distributions shift rightwards, yielding the intuitive result that richer spoofed point clouds produce more convincing phantom signatures. 
\vspace{-1.1em}
\paragraph{Finding 3.} \label{para:finding3} Figure~\ref{fig:attack_lifetime_removal} reveals that temporal detectors disproportionately rely on current-frame observations. Each subplot traces ASR across the full attack life cycle: attack onset (left of the dashed red line), where the current frame is spoofed and the contamination progressively fills the historical window, and attack decay (right of the dashed red line), where the current frame returns to clean and spoofed frames age out of the history. We consider CP-4f on the Removal 40 setting as an example.  ASR rises sharply from 0 ASR at $k=0$ (all clean) to \textasciitilde0.47 ASR at $k=1$ (current frame spoofed only), then increases marginally as additional historical frames are contaminated, before peaking at \textasciitilde0.9 ASR at $k=4$ (all four frames spoofed). At $k=5$, the current frame becomes clean while all historical frames remain spoofed, and ASR drops to \textasciitilde0.15, despite 75\% of the temporal window still containing evidence of spoofing. This asymmetry between attack onset and decay is consistent across all four detectors; a single spoofed current frame produces a larger ASR increase than all historical spoofed frames without a current spoofed frame. MSF-4f, MSF-8f, and PTT exhibit a similar pattern at their respective window sizes, with PTT sharply dropping at the transition (from ASR=\textasciitilde0.85 to \textasciitilde0.10 for Removal 40, for example) despite having 31 contaminated sets of proposals in its history. These results demonstrate that \textit{temporal detectors treat current-frame evidence as dominant, using historical context primarily for refinement} rather than as an equal contributor to detection decisions. 

\begin{table*}[t]
\centering
\caption{Point injection attack performance across spoofing modes and difficulty levels. Entries report Waymo L2 mAP$\uparrow$ and Attack Success Rate (ASR)$\downarrow$.}
\label{tab:injection_results}
\resizebox{\linewidth}{!}{
\normalsize
\begin{tabular}{lcc cc cc cc cc cc}
\toprule
& \multicolumn{2}{c}{Rel-Fixed Easy}
& \multicolumn{2}{c}{Rel-Fixed Medium}
& \multicolumn{2}{c}{Rel-Fixed Hard}
& \multicolumn{2}{c}{Global Easy}
& \multicolumn{2}{c}{Global Medium}
& \multicolumn{2}{c}{Global Hard} \\
\cmidrule(lr){2-3}
\cmidrule(lr){4-5}
\cmidrule(lr){6-7}
\cmidrule(lr){8-9}
\cmidrule(lr){10-11}
\cmidrule(lr){12-13}
Detector
& L2 mAP & ASR
& L2 mAP & ASR
& L2 mAP & ASR
& L2 mAP & ASR
& L2 mAP & ASR
& L2 mAP & ASR \\
\midrule
CenterPoint~\cite{CenterPoint}
& 65.5 & \underline{0.11}
& 65.1 & \textbf{0.12}
& 65.1 & \textbf{0.10}
& 65.5 & \underline{0.16}
& 65.3 & \underline{0.30}
& 65.2 & \textbf{0.32} \\

CenterFormer~\cite{CenterFormer}
& 68.9 & 0.12
& 68.7 & \underline{0.13}
& 68.6 & 0.20
& 68.9 & 0.18
& 68.7 & 0.31
& 68.7 & 0.34 \\

DSVT~\cite{DSVT}
& 72.1 & 0.12
& 71.8 & 0.21
& 71.7 & 0.28
& 72.1 & 0.18
& 71.9 & 0.33
& 71.8 & 0.37 \\

PHiM~\cite{phim}
& 68.3 & \textbf{0.07}
& 68.0 & 0.14
& 67.9 & 0.23
& 68.4 & \textbf{0.14}
& 68.2 & \textbf{0.26}
& 68.1 & \underline{0.33} \\

LoGoNet~\cite{logonet}
& 68.3 & \underline{0.11}
& 67.8 & \underline{0.13}
& 67.9 & \underline{0.12}
& 68.2 & 0.20
& 68.1 & 0.36
& 67.9 & 0.37 \\

CenterPoint-4F~\cite{CenterPoint}
& 68.8 & 0.12
& 68.5 & 0.22
& 68.5 & 0.29
& 68.9 & 0.18
& 68.8 & 0.36
& 68.8 & 0.45 \\

MSF-4f~\cite{He_2023_CVPR}
& \underline{74.6} & 0.13
& \underline{74.3} & 0.29
& \underline{74.3} & 0.35
& \underline{74.6} & 0.19
& \underline{74.5} & 0.41
& \underline{74.5} & 0.46 \\

MSF-8f~\cite{He_2023_CVPR}
& 74.2 & 0.14
& 73.9 & 0.31
& 73.8 & 0.39
& 74.3 & 0.23
& 74.2 & 0.49
& 74.1 & 0.54 \\

PTT~\cite{huang2024ptt}
& \textbf{75.0} & 0.14
& \textbf{74.7} & 0.26
& \textbf{74.7} & 0.35
& \textbf{75.0} & 0.21
& \textbf{74.9} & 0.44
& \textbf{74.9} & 0.50 \\
\bottomrule
\end{tabular}
}
\vspace{1pt}
{\footnotesize Best results are shown in \textbf{bold}, and second-best results are \underline{underlined}.
}
\vspace{-1em}
\end{table*}
\begin{table*}[t]
\centering
\caption{Point removal attack performance across removal configurations. Entries report Waymo L2 mAP$\uparrow$ and Attack Success Rate (ASR)$\downarrow$.}
\label{tab:removal_results}
\resizebox{\textwidth}{!}{
\begin{tabular}{lcc cc cc cc cc cc}
\toprule
& \multicolumn{2}{c}{Removal 10}
& \multicolumn{2}{c}{Removal 20}
& \multicolumn{2}{c}{Removal 30}
& \multicolumn{2}{c}{Removal 40}
& \multicolumn{2}{c}{Removal 50}
& \multicolumn{2}{c}{Removal 60} \\
\cmidrule(lr){2-3}
\cmidrule(lr){4-5}
\cmidrule(lr){6-7}
\cmidrule(lr){8-9}
\cmidrule(lr){10-11}
\cmidrule(lr){12-13}
Detector
& L2 mAP & ASR
& L2 mAP & ASR
& L2 mAP & ASR
& L2 mAP & ASR
& L2 mAP & ASR
& L2 mAP & ASR \\
\midrule
CenterPoint~\cite{CenterPoint}
& 64.3 & 0.90
& 62.4 & 1.00
& 60.0 & 0.96
& 60.7 & 0.95
& 60.7 & 0.77
& 60.7 & 0.77 \\

CenterFormer~\cite{CenterFormer}
& 67.9 & 0.85 
& 66.0 & 1.00 
& 65.3 & 0.94 
& 65.3 & 0.95 
& 65.2 & 0.74 
& 65.0 & 0.73 \\

DSVT~\cite{DSVT}
& 72.0 & 0.84
& 70.9 & 1.00
& 69.1 & 0.92
& 69.2 & 0.93
& 69.2 & 0.71
& 68.9 & 0.71 \\

PHiM~\cite{phim}
& 67.0 & 0.87
& 65.1 & \underline{0.99}
& 62.5 & 0.97
& 62.5 & 0.97
& 63.5 & 0.74
& 63.1 & 0.74 \\

LoGoNet~\cite{logonet}
& 66.9 & 0.84
& 64.9 & 1.00
& 62.6 & 0.93
& 62.6 & 0.93
& 63.9 & 0.68
& 63.6 & 0.68 \\

CenterPoint-4f~\cite{CenterPoint}
& 67.8 & 0.75
& 65.8 & \textbf{0.94}
& 63.9 & 0.83
& 66.0 & 0.84
& 66.1 & 0.60
& 66.1 & 0.59 \\

MSF-4f~\cite{He_2023_CVPR}
& \underline{73.4} & 0.73
& \underline{71.0} & 1.00
& \underline{69.2} & 0.83
& \underline{71.7} & 0.83
& \underline{71.7} & 0.54
& \underline{71.7} & \underline{0.53} \\

MSF-8f~\cite{He_2023_CVPR}
& 73.0 & \underline{0.72}
& 70.7 & \underline{0.99}
& 68.9 & \underline{0.82}
& 71.3 & \underline{0.82}
& 71.4 & \underline{0.54}
& 71.4 & 0.54 \\

PTT~\cite{huang2024ptt}
& \textbf{73.8} & \textbf{0.68}
& \textbf{71.4} & \underline{0.99}
& \textbf{69.6} & \textbf{0.79}
& \textbf{72.2} & \textbf{0.79}
& \textbf{72.2} & \textbf{0.51}
& \textbf{72.3} & \textbf{0.50} \\
\bottomrule
\end{tabular}
}
\vspace{1pt}
{\footnotesize Best results are shown in \textbf{bold}, and second-best results are \underline{underlined}.
}
\vspace{-2em}
\end{table*}

\vspace{-1em}
\section{Discussion}\label{sec:discussion}
\vspace{-1em}
In this section, we examine our most counterintuitive finding: models that perform better on clean benchmarks are often more vulnerable to point-injection attacks than weaker counterparts. We consider two possible explanations. \textbf{First}, memory mechanisms may amplify small corruptions: once injected geometry is incorporated into a temporal state, subsequent predictions may inherit or compound the error. \textbf{Second}, standard training augmentations \cite{openpcdet, MMDetection3D, zhu2019class} may inadvertently resemble the perturbations used by point-injection attacks, causing spoofed geometry to appear in-distribution.

These hypotheses are motivated by two observations. Existing multi-frame detectors typically incorporate memory late in the detection pipeline, after proposals or detections have already been produced. As a result, the current observation is often merged into memory with limited scrutiny, effectively granting the present frame full trust. This design makes robustness strongly dependent on the reliability of the instantaneous observation, consistent with Fig.~\ref{fig:attack_lifetime_removal} and Fig.~\ref{fig:attack_lifetime_injection}. Separately, many LiDAR detectors are trained with ground-truth sampling, where object traces are copied from training scenes and pasted into other point clouds to increase foreground density. Although effective for clean benchmark performance, this procedure can violate the physical constraints of real LiDAR scans, such as occlusion, visibility, and sensor-return consistency. In this sense, ground-truth sampling is structurally similar to point injection. Models that excel under such training may therefore learn to treat inserted object-like point patterns as plausible scene content, increasing their susceptibility to spoofed objects at test time.

To isolate these mechanisms, we introduce \emph{Latent Occupancy Tracking} (LOT), a lightweight memory module that operates at the beginning of the detection pipeline. LOT maintains an online latent expectation of scene occupancy and compares it against the current observation before downstream detection, allowing the model to estimate uncertainty prior to proposal generation. We deploy LOT on PHiM because its streaming state-space formulation over partial sectors provides a controlled substrate for studying early-stage temporal context: unlike conventional multi-frame detectors, PHiM does not rely on late proposal- or detection-level fusion, allowing us to evaluate whether latent memory itself improves robustness without conflating it with downstream aggregation effects. Additional implementation details are provided in Appendix~\ref{app:lot}.

\begin{table*}[t]
\centering
\caption{Effect of LOT on robustness under different point injection attacks.}
\label{tab:LOT_ablation_injection}
\resizebox{\linewidth}{!}{
\normalsize
\begin{tabular}{l c cc cc cc cc cc cc}
\toprule
& Clean
& \multicolumn{2}{c}{Rel-Fixed Easy}
& \multicolumn{2}{c}{Rel-Fixed Medium}
& \multicolumn{2}{c}{Rel-Fixed Hard}
& \multicolumn{2}{c}{Global Easy}
& \multicolumn{2}{c}{Global Medium}
& \multicolumn{2}{c}{Global Hard} \\
\cmidrule(lr){2-2}
\cmidrule(lr){3-4}
\cmidrule(lr){5-6}
\cmidrule(lr){7-8}
\cmidrule(lr){9-10}
\cmidrule(lr){11-12}
\cmidrule(lr){13-14}
Detector
& L2 mAP
& L2 mAP & ASR
& L2 mAP & ASR
& L2 mAP & ASR
& L2 mAP & ASR
& L2 mAP & ASR
& L2 mAP & ASR \\
\midrule

PHiM~\cite{phim}
& 48.97
& 48.26 & 0.0642
& 48.26 & 0.1888
& 48.02 & 0.2817
& 48.40 & 0.0918
& 48.22 & 0.2442
& 48.04 & 0.3174 \\

PHiM + LOT
& 54.28
& 53.45 & 0.0985
& 53.21 & 0.2089
& 55.08 & 0.2753
& 53.44 & 0.1222
& 53.42 & 0.3016
& 53.20 & 0.3233 \\

\bottomrule
\end{tabular}
}

\vspace{1pt}
{\footnotesize Models are trained on the 1/20 split of the Waymo dataset and validated on 1/5 adversarial dataset.}

\end{table*}
\begin{table*}[!t]
\vspace{-1em}
\centering
\caption{Effect of LOT on robustness under point removal attacks across different removal levels.}
\label{tab:LOT_ablation_removal}
\resizebox{\textwidth}{!}{
\begin{tabular}{l c cc cc cc cc cc cc}
\toprule
& Clean
& \multicolumn{2}{c}{Removal 10}
& \multicolumn{2}{c}{Removal 20}
& \multicolumn{2}{c}{Removal 30}
& \multicolumn{2}{c}{Removal 40}
& \multicolumn{2}{c}{Removal 50}
& \multicolumn{2}{c}{Removal 60} \\
\cmidrule(lr){2-2}
\cmidrule(lr){3-4}
\cmidrule(lr){5-6}
\cmidrule(lr){7-8}
\cmidrule(lr){9-10}
\cmidrule(lr){11-12}
\cmidrule(lr){13-14}
Detector
& L2 mAP
& L2 mAP & ASR
& L2 mAP & ASR
& L2 mAP & ASR
& L2 mAP & ASR
& L2 mAP & ASR
& L2 mAP & ASR \\
\midrule

PHiM~\cite{phim}
& 48.97
& 47.25 & 0.9502
& 46.06 & 0.9990
& 44.13 & 0.9812
& 44.22 & 0.9828
& 44.39 & 0.8622
& 43.90 & 0.8680 \\

PHiM + LOT
& 54.28
& 52.45 & 0.9296
& 51.05 & 0.9987
& 48.83 & 0.9780
& 49.07 & 0.9771
& 49.42 & 0.8192
& 48.99 & 0.8225 \\

\bottomrule
\end{tabular}
}

\vspace{1pt}
{\footnotesize Models are trained on the 1/20 split of the Waymo dataset and validated on 1/5 adversarial dataset.}
\vspace{-2em}
\end{table*}

Under point-removal attacks (Tab. \ref{tab:LOT_ablation_removal}), LOT consistently improves robustness, suggesting that sequential memory can help recover missing geometry from contextual occupancy expectations even in a single-frame detection setting. In contrast, under point-injection attacks (Tab. \ref{tab:LOT_ablation_injection}), LOT increases ASR by approximately 3 percentage points relative to base PHiM. Because the attack is confined to a single frame and LOT filters the input before downstream detection, this degradation is unlikely to arise from temporal error accumulation alone. Instead, it supports the second hypothesis: training with ground-truth sampling may bias stronger or more context-sensitive models toward accepting inserted object-like structures as plausible. Thus, while early latent memory can mitigate attacks that remove evidence, it does not necessarily protect against attacks that add plausible but spurious evidence.
\vspace{-1em}
\section{Conclusion}
\vspace{-1em}
We introduce ATLAS, a large-scale benchmark for evaluating black-box adversarial robustness of LiDAR perception  under physically grounded point injection and removal attacks. Across nine detectors spanning single-frame, multi-frame, streaming, and camera-LiDAR fusion architectures, ATLAS reveals a robustness asymmetry: stronger models better withstand point removal attacks but become more vulnerable to point injection attacks. We trace this to late-stage temporal aggregation granting current frames disproportionate trust, and to ground-truth sampling augmentation structurally resembling injection attacks. While ATLAS simulates the attacks physics via ray casting and empirical removal rates from prior works, validating these findings against hardware spoofing rigs remains an important future work to confirm that our architectural rankings hold under physical spoofing attacks. Our LOT case study shows early latent memory can mitigate the effects of point removal but not injection, suggesting training-time interventions are necessary alongside architectural defenses. 
We release ATLAS to make black-box sensor robustness a measurable axis of perception development as physical attack capabilities evolve. 

\newpage

{\small
\bibliographystyle{plain}
\bibliography{main}
}


\appendix
\clearpage
\setcounter{page}{1}
\maketitle

\section{Dataset Generation Algorithms} 

\label{app:algs}
\begin{algorithm}[H]
\caption{Phantom point injection dataset generation}
\label{alg:injection}
\begin{algorithmic}[1]
\Require Point cloud sequence $\{\mathcal{P}_t\}_{t=1}^{T}$,
         ego-pose sequence $\{\mathbf{T}_t \in SE(3)\}_{t=1}^{T}$,
         trace library $\mathcal{L}$,
         perturbation mode $m \in \{\textsc{Global}, \textsc{Relative}\}$,
         seed $s$
\Ensure  Spoofed sequence $\{\tilde{\mathcal{P}}_t\}$,
         phantom GT boxes $\{\tilde{\mathbf{b}}_t\}$

\State $\mathcal{A} \gets \textsc{ToggleSchedule}(\{1\ldots T\},\, w{=}32)$
\State $\mathcal{R} \gets \textsc{ExtractRuns}(\mathcal{A})$

\For{each run $\mathcal{W} = [t_s, t_e] \in \mathcal{R}$}
    \State $\tilde{\mathbf{p}},\,\tilde{\mathbf{b}} \gets \textsc{SampleTrace}(\mathcal{L},\, s{+}t_s)$
    \State $\tilde{\mathbf{p}} \gets \tilde{\mathbf{p}} - \mathrm{mean}(\tilde{\mathbf{p}}[:,:2])$
    \State $\tilde{\mathbf{p}} \gets \mathbf{R}(\psi)\,\tilde{\mathbf{p}}$
    \Comment{offset $\delta \sim \mathcal{U}([15,20]\times[-3,3])$, $\psi \sim \mathcal{U}(-0.2,0.2)$}

    \If{$m = \textsc{Global}$}
        \State $\mathbf{T}_{\text{anc}} \gets \mathbf{T}_{t_e}$
        \State $\tilde{\mathbf{p}}^{W} \gets \mathbf{T}_{\text{anc}}\,\tilde{\mathbf{p}}$
        \State $\tilde{\mathbf{b}}^{W} \gets \mathbf{T}_{\text{anc}}\,\tilde{\mathbf{b}}$
    \EndIf

    \For{each frame $t \in \mathcal{W}$}
        \If{$m = \textsc{Global}$}
            \State $\tilde{\mathbf{p}}_t \gets \mathbf{T}_t^{-1}\,\tilde{\mathbf{p}}^{W}$
            \State $\tilde{\mathbf{b}}_t \gets \mathbf{T}_t^{-1}\,\tilde{\mathbf{b}}^{W}$
        \Else
            \State $\tilde{\mathbf{p}}_t \gets \tilde{\mathbf{p}}$
            \State $\tilde{\mathbf{b}}_t \gets \tilde{\mathbf{b}}$
        \EndIf

        \State $\tilde{\mathbf{p}}_t \gets \textsc{GroundAnchor}(\tilde{\mathbf{p}}_t,\, \mathcal{P}_t)$
        \State $\tilde{\mathcal{P}}_t \gets \textsc{Raycast}(\mathcal{P}_t,\, \tilde{\mathbf{p}}_t)$
    \EndFor
\EndFor
\end{algorithmic}
\end{algorithm}

\begin{algorithm}[H]
\caption{Point removal dataset generation}
\label{alg:removal}
\begin{algorithmic}[1]
\small
\Require Point cloud sequence $\{\mathcal{P}_t\}_{t=1}^{T}$,
         GT box sequence $\{\mathcal{B}_t\}$,
         azimuth width $\alpha$,
         elevation width $\beta = 16°$,
         seed $s$
\Ensure  Attacked sequence $\{\tilde{\mathcal{P}}_t\}$

\State $\mathcal{A} \gets \textsc{ToggleSchedule}(\{1\ldots T\},\, w{=}32)$
\State $\mathcal{R} \gets \textsc{ExtractRuns}(\mathcal{A})$

\For{each run $\mathcal{W} = [t_s, t_e] \in \mathcal{R}$}
    \State Select initial target box $\mathbf{b}_{t_s}$ from $\mathcal{B}_{t_s}$
    
    \For{each frame $t \in \mathcal{W}$}
        \State $\hat{\mathbf{c}}_t \gets \mathbf{c}_{t-1} + \Delta t [v_x,\, v_y,\, 0]^\top$ 
        \Comment{predict next pos.}
        
        \State $\mathbf{b}_t \gets$ nearest same-class box to $\hat{\mathbf{c}}_t$ with displacement $< d_{\max}$
        
        \If{$\mathbf{b}_t = \varnothing$} \Comment{tracking lost}
            \State $\tilde{\mathcal{P}}_t \gets \mathcal{P}_t$;\quad \textbf{continue}
        \EndIf
        
        \State Compute $(\theta^*_t,\, \phi^*_t)$ from $\mathbf{b}_t$ center
        
        \State $a_t \sim \mathrm{Bernoulli}(p_{\text{dist}}(r_t))$ 
        \Comment{firing gate}
        
        \If{$a_t = 0$}
            \State $\tilde{\mathcal{P}}_t \gets \mathcal{P}_t$;\quad \textbf{continue}
        \EndIf
        
        \State $\mathcal{S}_t \gets \{i : |\mathrm{wrap}(\theta_i - \theta^*_t)| \leq \tfrac{\alpha}{2},\;
               |\phi_i - \phi^*_t| \leq \tfrac{\beta}{2}\}$ 
        
        \State $\mathcal{M}_t \sim \mathrm{Bernoulli}(p_{\text{remove}})^{|\mathcal{S}_t|}$
        
        \State $\tilde{\mathcal{P}}_t \gets \mathcal{P}_t \setminus \{\mathbf{p}_i : i \in \mathcal{S}_t,\, \mathcal{M}_{t,i} = 1\}$
    \EndFor
\EndFor
\end{algorithmic}
\end{algorithm}

\section{Baselines}
\label{app:baselines}
\noindent \textbf{CenterPoint.} CenterPoint \cite{CenterPoint} encodes a LiDAR point cloud into a bird's-eye-view (BEV) feature map using a 3D backbone, then applies convolution-based heads to produce a class-specific heatmap whose local peaks identify object centers. From each center, the network regresses 3D size, orientation, and velocity, unifying detection and tracking in a single anchor-free framework. A second-stage refinement extracts bilinear-interpolated face-center features from predicted bounding boxes to produce IoU-aware confidence scores and refined box estimates.
CenterPoint 4-frame (referred to as CP4F) extends CenterPoint by concatenating four consecutive LiDAR sweeps into a single point cloud before encoding. This multi-frame aggregation produces denser voxel evidence, yielding richer representations particularly for stationary objects that accumulate points across frames.

\noindent \textbf{MSF.\quad} MSF \cite{He_2023_CVPR} is a two-stage temporal detector that uses CenterPoint 4-frame as its region proposal network (RPN). Current-frame proposals are propagated backward using estimated velocities,  and 128 raw LiDAR points from each historical frame are pooled within expanding cylindrical regions centered on the propagated proposal locations. The sampled points are encoded via geometric and motion embeddings, then refined through three learning blocks containing multi-head self-attention and a Bidirectional Feature Aggregation (BiFA) module. BiFA enables cross-frame information exchange in both temporal directions, allowing persistent patterns across frames to be mutually reinforced. A query-based transformer decoder produces the final proposal representation for box regression and confidence prediction. In this work, we evaluate the performance of MSF with a four-frame and eight-frame point cloud histories.

\noindent \textbf{PTT.\quad} PTT \cite{huang2024ptt} is a two-stage temporal detector that uses CP4F as its RPN. It requires only the current frame's point cloud alongside a 32 frame history of proposals, avoiding storage of historical point clouds. Point-to-Proposal features encode geometric relationships between current-frame points and historical proposal box corners, while Proposal-to-Proposal features capture inter-frame displacements. These are combined into per-frame point-trajectory features. These features are split into long-term and short-term memories, where short-term features query the long-term memory via cross-attention to extract relevant historical trends. A future-aware module further encodes anticipated trajectory features from estimated velocities. Finally, a point-trajectory aggregator fuses all memory outputs with current-frame point features to produce final detections.

\noindent \textbf{LoGoNet. \quad} LoGoNet \cite{logonet} is a LiDAR-camera fusion detector designed for Waymo-scale 3D object detection. It fuses the two modalities at both global and local scales: a Global Fusion module aligns image features with LiDAR voxel features across the full scene using point centroids for accurate cross-modal projection, while a Local Fusion module subdivides each 3D proposal into a uniform grid, projects grid centers into camera images, and fuses sampled image features with position-decorated point-cloud features for each candidate region. A Feature Dynamic Aggregation module then allows locally and globally fused representations to interact before final box refinement. Because LoGoNet is publicly available and natively supports Waymo, no custom re-implementation was required. From an adversarial robustness perspective, LoGoNet is not explicitly designed as a defense. Its fusion strategy primarily enhances LiDAR features with image cues rather than performing explicit cross-modal consistency checks. As a result, a LiDAR-only phantom that successfully induces a strong geometric response may still persist if the fused representation remains confident, even in the absence of clear supporting evidence from the camera. LoGoNet is evaluated against all six point injection modes and all six point removal widths (10\textdegree--60\textdegree).

\noindent \textbf{PHiM. \quad} PHiM \cite{phim} is a State Space Model (SSM) architecture designed specifically for polar-coordinate streaming LiDAR object detection. Instead of relying on traditional 3D convolutions, PHiM leverages a custom Mamba-based streaming backbone with dimensionally-decomposed operations to avoid the heavy distortions typical of polar representations. It employs local bidirectional Mamba blocks for intra-sector spatial encoding and a global forward Mamba block for inter-sector temporal modeling. Local sector features are extracted via this backbone and accumulated into a sector feature buffer, enabling efficient inter-sector communication through a full-scan backbone. This hybrid approach combines the rapid update rates of polar processing with accurate full-scene reasoning, processing point clouds as egocentric sequences to overcome the limited visibility and cross-sector dependencies of prior streaming methods.

\noindent \textbf{DSVT. \quad} DSVT \cite{DSVT} is a single-stride, window-based voxel Transformer backbone designed to efficiently handle the inherent sparsity of outdoor point clouds. To process sparse voxels in a fully parallel manner without relying on customized CUDA operations, DSVT introduces Dynamic Sparse Window Attention, which partitions a series of local regions within each window according to their sparsity. To enable cross-set feature exchange, it employs a rotated set partitioning strategy that alternates between two spatial partitioning configurations across consecutive self-attention layers. Additionally, DSVT utilizes an attention-style 3D pooling module to downsample features while effectively encoding geometric information. This purely Transformer-based architecture avoids the representational limitations of sparse convolutions and serves as a highly deployment-friendly, general-purpose 3D perception backbone.

\noindent \textbf{CenterFormer. \quad} CenterFormer \cite{CenterFormer} is a center-based transformer network that constructs long-range feature attention for 3D object detection while mitigating the immense computational complexity typical of standard query-based transformers on large point clouds. The architecture first utilizes a standard voxel-based backbone and a center heatmap to identify a limited number of object center candidates. The features extracted from these center candidates are then explicitly used as the initial query embeddings in the transformer module. A cross-attention mechanism is employed to aggregate contextual features from neighboring spatial regions around each candidate. To leverage temporal context, CenterFormer further incorporates a multi-frame cross-attention mechanism that aligns and fuses object features from historical frames, allowing the network to robustly handle moving objects before applying regression heads to produce final bounding box predictions.

\section{Evaluation Implementation} \label{app:eval_implementation}
\noindent\textbf{Single frame method evaluation.} Because ATLAS consists of spoofed single-frames, evaluating single-frame methods \cite{CenterFormer, CenterPoint, phim, DSVT}
\noindent\textbf{Proposal generation with CenterPoint.}
MSF \cite{He_2023_CVPR} and PTT \cite{huang2024ptt} use CP4F as a first stage. To ensure that spoofed observations propagate through the detection pipeline realistically, we recompute region proposals using CP4F \cite{CenterPoint}. 

First, spoofed points are ray-casted into the four-frame history of a current frame based on the spoof schedule described by Eq. \eqref{eq:schedule}. Then, for each frame $t$, a four-frame point cloud is constructed by transforming the single-frame point clouds from frames $t-3, ..., t$ into the ego coordinate frame at time $t$. This can be represented by Eq. \eqref{eq:4frame_agg},
\begin{equation}
    \mathcal{P}^{(t)} = \bigcup_{j=0}^{3} \left\{ \left[\mathbf{T}_t^{-1} \mathbf{T}_{t-j} \tilde{\mathbf{p}},\; p_{3:5},\; 0.1j \right] : \mathbf{p} \in \mathcal{P}_{t-j} \right\},
    \label{eq:4frame_agg}
\end{equation}
where $\mathbf{T}_k \in SE(3)$ is the ego-to-global transformation at frame $k$, $\tilde{\mathbf{p}} = [p_x, p_y, p_z, 1]^\top$ is the homogeneous point coordinate, $p_{3:5}$ are the intensity and elongation channels, and $0.1j$ is a lag timestamp encoding the frame's temporal offset in seconds. When all four frames in the history are clean,  the original unmodified dataloader points are used; for frames containing spoofed objects in the history, the modified aggregated cloud is constructed. A CenterPoint forward pass is performed, and the resulting proposals, labels, and confidence scores are stored per-frame for use as RPN input to the temporal detectors.

\noindent\textbf{MSF Evaluation.} For each frame $t$, the four-frame point cloud is reconstructed from the spoofed dataset following Eq. \eqref{eq:4frame_agg}. A modified version of Eq. \eqref{eq:4frame_agg} which goes up to i$j=7$ is used to evaluate the eight-frame variant of MSF. The stored CenterPoint proposals for the current frame undergo a velocity rescaling identical to the one used by the official MSF pipeline, as shown in Eq. \eqref{eq:vel_rescale}:
\begin{equation}
    (\hat{v}_{x}, \hat{v}_{y}) = (-0.1 \cdot v_x,\; -0.1 \cdot v_y).
    \label{eq:vel_rescale}
\end{equation}
This converts the CenterPoint proposal velocities into backward motion used by MSF's motion-guided point sampler to propagate proposals to their estimated past locations. The transformed bounding box proposals, scores, labels, and four/eight-frame point cloud are used for an MSF forward pass. 

 \noindent\textbf{PTT evaluation.} For each frame $t$, the stored proposals from the preceding 32 frames are collected, velocity-rescaled via Eq. \eqref{eq:vel_rescale}, and transformed into the ego coordinate frame of frame $t$:
\begin{equation}
    \hat{\mathbf{B}}_k = f_{\mathrm{box}}\!\left(\mathbf{T}_t^{-1}\mathbf{T}_k,\; \mathbf{B}_k\right), \quad k = t-31, \ldots, t,
    \label{eq:ptt_transform}
\end{equation}
where $f_{\text{box}}$ denotes the rigid-body transformation of box center, heading, and velocity. A PTT forward pass is then performed, using the current frame's point cloud and transformed proposal history, label history, and confidence score history as inputs.

\section{Extended Results} \label{app:extendedresults}
\begin{figure}[htbp]
    \centering
    \includegraphics[width=\linewidth]{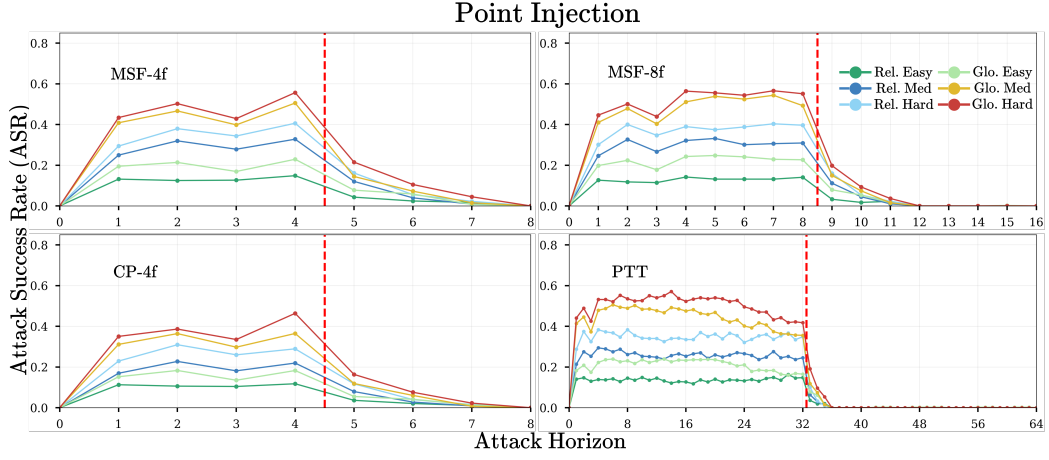}
    \caption{\textbf{Decay of injection vulnerability.} The ASR peaks at the end of the attack frames before decaying monotonically as clean frames enter the temporal memory buffer.}
    \label{fig:attack_lifetime_injection}
\end{figure}

Figure \ref{fig:attack_lifetime_injection} above illustrates the ASR in a chronological timeline of frames, highlighting how the temporal vulnerability peaks immediately before the injection attack ends. Once the continuous introduction of spoofed points ends, the memory buffer begins processing the actual spatial data instead of adversarial features. As a result, the ASR exhibits rapid and monotonic decay as incoming clean frames successfully dull the compromised history of the system. 

\begin{figure}[htbp]
    \centering
    \includegraphics[width=\linewidth]{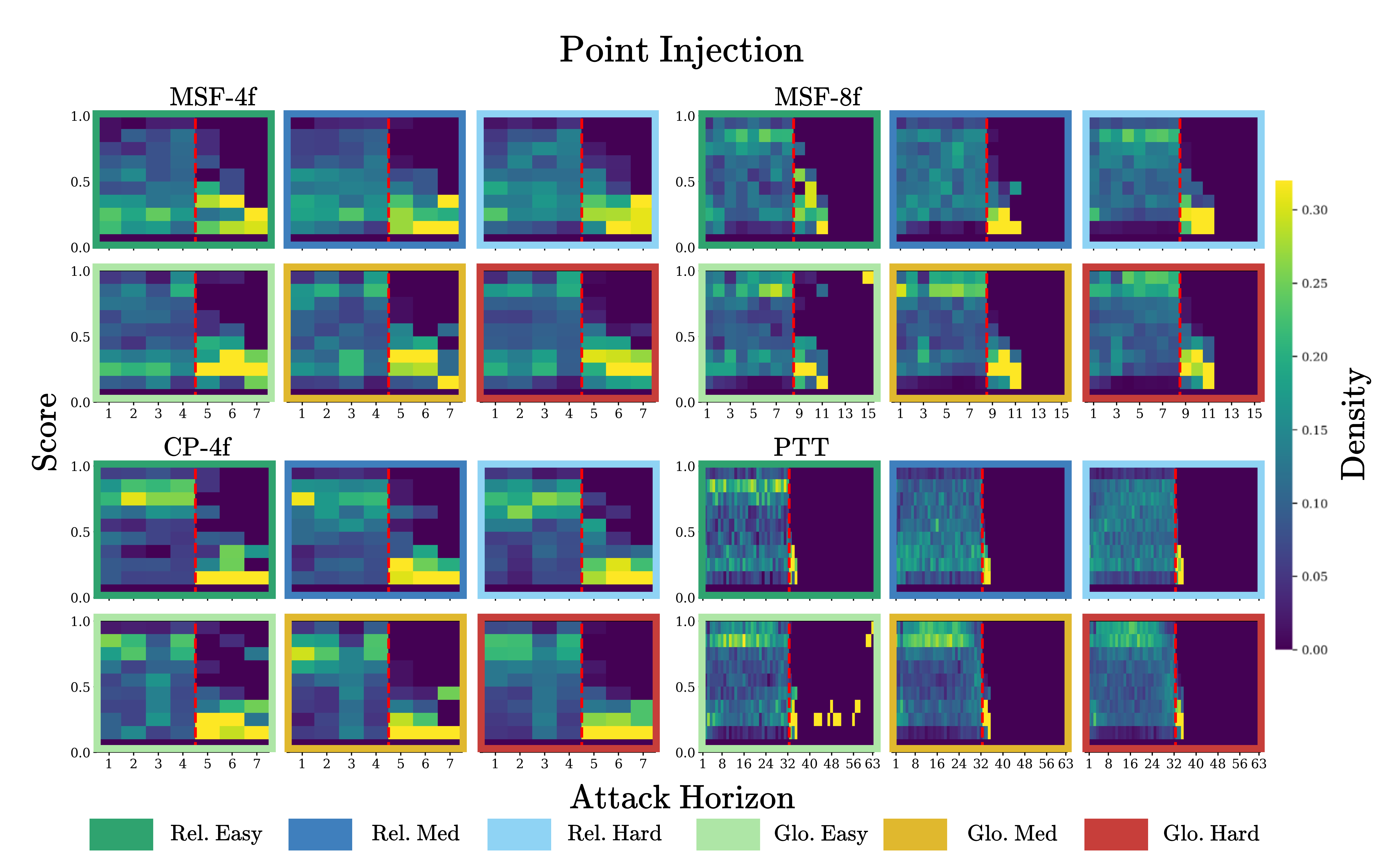}
    \caption{\textbf{False confidence during point injection.} Detectors assign high confidence to injected phantom objects during the active attack frames, which rapidly dissipates once the temporal buffer is overwritten with uncompromised frames.}    \label{fig:injection_heatmap}
\end{figure}
As shown in Figure \ref{fig:injection_heatmap} above, the confidence score distributions for point injection attacks skew toward maximum certainty during the active spoofing phase because these dense and high-confidence clusters confirm that temporal detectors fail to reject phantoms and instead actively reinforce them until the attack threshold is crossed. During the following recovery phase, this false confidence rapidly decreases as uncompromised frames enter the temporal window and overwrite the spoofed history.  

\begin{figure}[htbp]
    \centering
    \includegraphics[width=\textwidth]{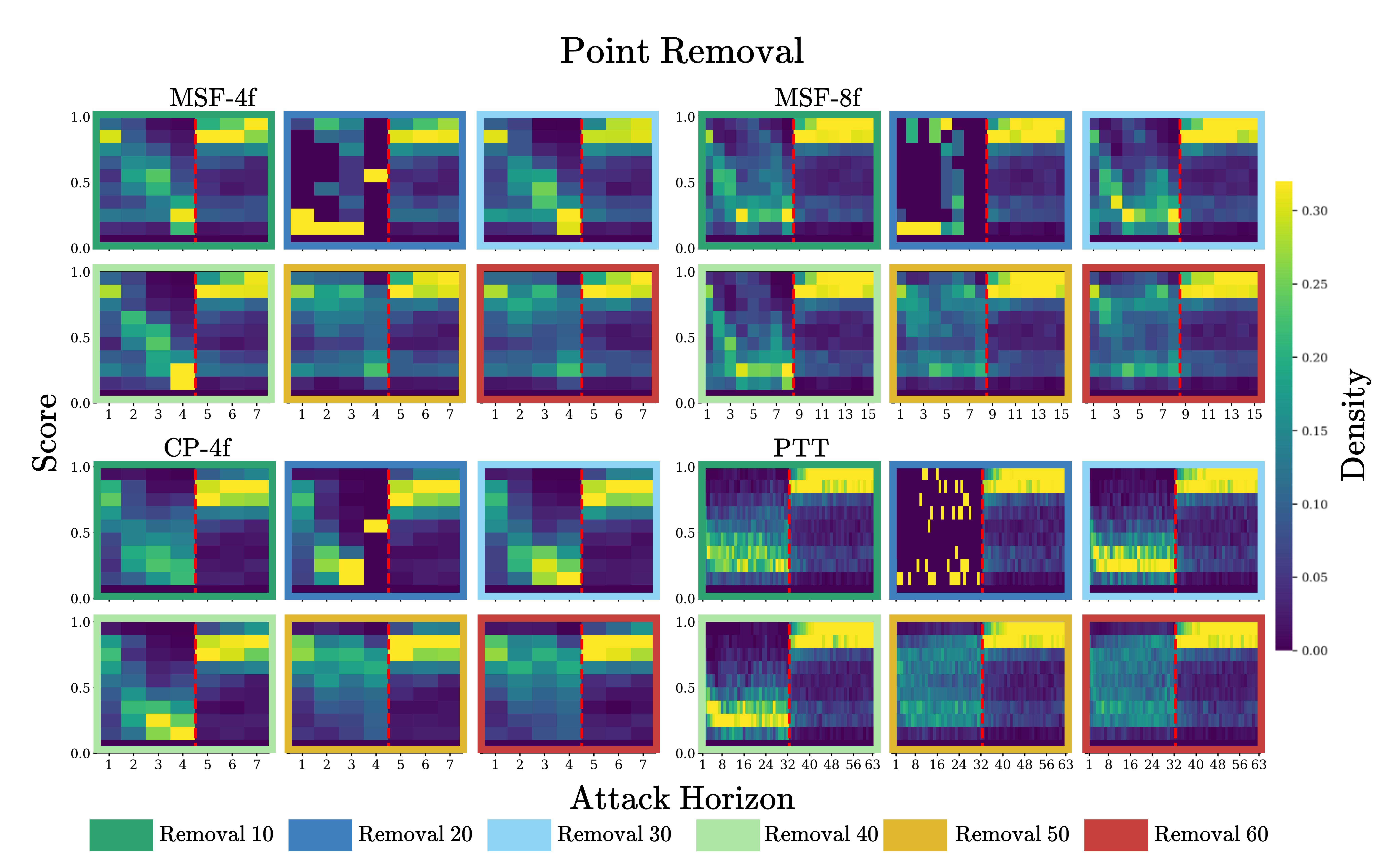}
    \caption{\textbf{Target confidence degradation and recovery.} Point removal systematically suppresses real target confidence scores during the active attack, followed by a full recovery to baseline certainty as clean point clouds reconstruct the scene.}    
    \label{fig:removal_heatmap}
\end{figure}
In contrast, Figure \ref{fig:removal_heatmap} demonstrates the shifting confidence distributions of real targets subjected to point removal across the temporal history buffer. During the active attack frames, the system's spatial perception is significantly disrupted, causing detectable target objects to exhibit significantly degraded confidence scores relative to their respective clean baselines. Once the adversarial removal frames conclude, the continuous flow of clean point clouds enables  temporal fusion modules to reconstruct accurate object representations and recover baseline confidence scores.

\begin{figure}[htbp]
    \centering
    \includegraphics[width=\textwidth]{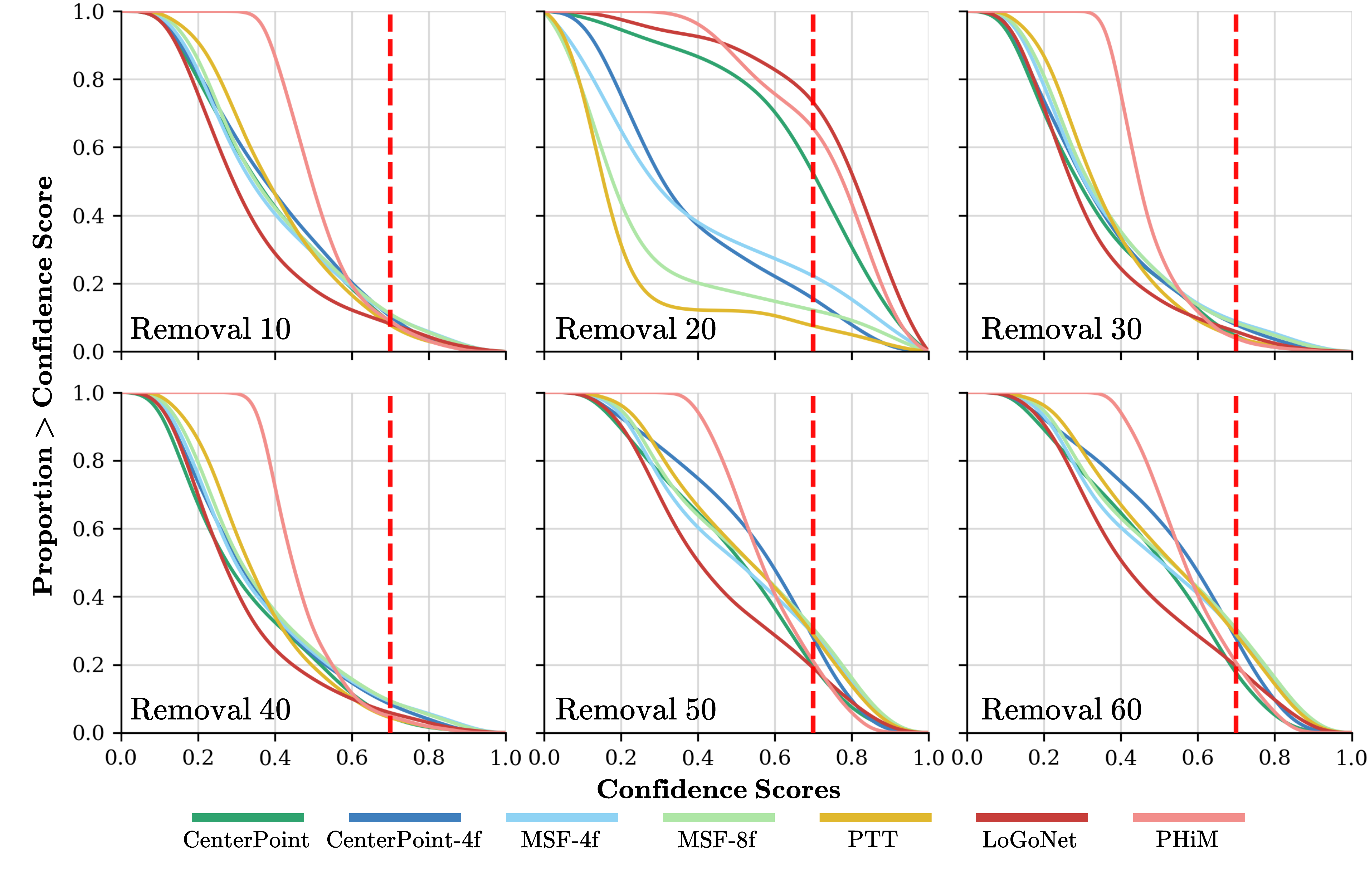}
    \caption{\textbf{Systematic degradation of target confidence in removal attacks.} Adversarial point removal effectively suppresses the certainty of legitimate detections, driving a significant proportion of target confidence scores below the 0.7 threshold.}    
    \label{fig:confidence_removal}
\end{figure}
\vspace{5mm}

\begin{table*}[h!]
\vspace{-2em}
\centering
\caption{Center gap between clean and perturbed score peaks across point removal levels.}
\label{tab:center_gap}
\resizebox{\linewidth}{!}{
\begin{tabular}{lcccccc}
\toprule
Method & Removal 10 & Removal 20 & Removal 30 & Removal 40 & Removal 50 & Removal 60 \\
\midrule
CenterPoint~\cite{CenterPoint} & 0.614 & 0.158 & 0.724 & 0.722 & 0.168 & 0.148 \\
CenterPoint 4F~\cite{CenterPoint} & 0.581 & 0.622 & 0.674 & 0.664 & 0.123 & 0.130 \\
MSF4~\cite{He_2023_CVPR} & 0.657 & 0.742 & 0.672 & 0.682 & 0.617 & 0.614 \\
MSF8~\cite{He_2023_CVPR} & 0.659 & 0.759 & 0.672 & 0.679 & 0.622 & 0.627 \\
PTT~\cite{huang2024ptt} & 0.594 & 0.759 & 0.637 & 0.634 & 0.586 & 0.586 \\
LoGoNet~\cite{logonet} & 0.682 & 0.063 & 0.707 & 0.722 & 0.604 & 0.602 \\
PHiM~\cite{phim} & 0.363 & 0.058 & 0.451 & 0.464 & 0.328 & 0.308 \\
\bottomrule
\end{tabular}
}
\end{table*}
Figure \ref{fig:confidence_removal} shows the score distribution of objects that the removal attack failed to hide. For most target removal azimuths, the proportion of surviving detection scores above 0.7 is less than $20\%$. Moreover, Table \ref{tab:center_gap} illustrates the difference in peaks of the object's spoofed score distribution and the peak of the same object's clean score distribution. Combined with the result seen in Fig. \ref{fig:confidence_removal}, we observe that the majority of detections that survive point removal have much lower confidence scores than the same detection in a clean setting. This yields the expected conclusion that point removal-attacked detections that survive the IoU threshold have heavily degraded scores.

\vspace{3mm}
\section{LOT Implementation Details}\label{app:lot}

Our proposed Latent Occupancy Tracking (LOT) module operates in the polar occupancy domain to model temporal consistency of LiDAR observations under adversarial perturbations. The implementation maintains a latent occupancy representation over discretized polar coordinates, where each voxel is indexed by $(z,t,r)$ corresponding to height, azimuth, and radial distance bins, respectively.

\subsection{Polar Occupancy Construction}

Given voxelized point features extracted by the backbone, LOT first predicts a scalar occupancy confidence for each voxel using a lightweight MLP which consists of two linear layers with GELU activation in between.

The predicted scores are scattered into a dense polar occupancy grid using the corresponding voxel coordinates. 

To reduce computational complexity, the occupancy map is collapsed along the height dimension through masked averaging:
\[
\mathcal{O}_{2D}(t,r)
=
\frac{\sum_z \mathcal{O}(z,t,r)\cdot M(z,t,r)}
{\sum_z M(z,t,r)},
\]
where $M$ denotes the voxel validity mask. The resulting 2D occupancy map is further downsampled using average pooling over the angular and radial dimensions. 

The full azimuth space is partitioned into multiple sectors, and each sector is processed independently. A fixed-length buffer stores historical sector occupancy maps for forecasting. 

\subsection{Forecasting via ConvSSM}

LOT employs a convolutional state-space model (ConvSSM) to predict future occupancy states from historical occupancy observations. Given a sequence of historical occupancy maps
\[
\{X_{t-K+1}, \dots, X_t\},
\]
the hidden state evolves according to:
\[
h_k = A h_{k-1} + B(X_k),
\]
where $A$ is a learnable diagonal state transition matrix parameterized in log-space for stability, and $B(\cdot)$ is implemented using a convolutional projection layer. 

The predicted occupancy map is decoded as:
\[
\hat{X}_{t+1} = C(h_t) + D X_t,
\]
where $C(\cdot)$ is a convolutional decoder and $D$ denotes a learnable skip connection from the latest observation. 

The final latent occupancy representation is obtained by merging the current observation with the predicted future occupancy:
\[
X^{\mathrm{final}}
=
0.5 X^{\mathrm{current}}
+
0.5 X^{\mathrm{forecast}}.
\]

To suppress adversarial or inconsistent voxels, LOT computes voxel-wise confidence scores from the predicted occupancy map and dynamically filters low-confidence voxels before passing features to downstream detection heads.

\subsection{Ground-Truth Occupancy Generation}

Ground-truth occupancy supervision is generated directly from annotated bounding boxes in polar coordinates. Each box is projected onto the polar grid according to its radial and angular extent, producing a binary occupancy mask. 

The occupancy map is subsequently downsampled and divided into sectors. To provide smoother supervision, Gaussian kernels are applied around occupied cells:
\[
G(t,r)
=
\exp\left(
-
\frac{(t-t_c)^2}{2\sigma_t^2}
-
\frac{(r-r_c)^2}{2\sigma_r^2}
\right).
\]

The final supervision map is obtained by combining the Gaussian occupancy target with the binary sector occupancy map.

\subsection{Training Strategy}

Training is performed in a staged manner to stabilize optimization and disentangle occupancy estimation, temporal forecasting, and downstream detection learning.

\paragraph{Stage 1: Score MLP Pretraining.}
In the first stage, only the voxel scoring network (\texttt{score\_mlp}) is trained. The objective is to learn reliable occupancy estimation from spatial voxel features before introducing SSM modeling.

\paragraph{Stage 2: SSM Training.}
After the occupancy scoring network converges, the ConvSSM module is optimized while keeping the pretrained scoring network fixed. Since the forecasting input consists of different spatial sectors within the same LiDAR frame , the SSM is trained to model spatial occupancy continuity and sector-wise latent consistency across neighboring regions.

The ConvSSM branch is supervised using a CenterNet-style focal occupancy loss:
\[
\mathcal{L}_{forecast}
=
\mathcal{L}_{focal}(\hat{X}, X^{gt}),
\]
where $\hat{X}$ denotes the predicted occupancy map and $X^{gt}$ represents the generated occupancy supervision target.

\paragraph{Stage 3: Detector Fine-Tuning.}
Finally, the entire VFE and LOT modules are frozen, including both the occupancy scoring network and ConvSSM branch. The remaining detector components are then fine-tuned using the filtered latent occupancy features for downstream 3D object detection.

\section{Broader impact}
As advances are made in attack hardware, the threat of autonomous vehicles operating in adversarial sensing conditions continues to grow. We envision ATLAS helping drive the development and evaluation of the next generation of LiDAR object detectors by enabling rigorous assessment of their adversarial robustness capabilities.


\end{document}